\documentclass[]{interact}

\usepackage{epstopdf}
\usepackage{subfigure}

\usepackage{graphicx}
\graphicspath{ {./figures/} }
\usepackage{lineno}
\usepackage[colorlinks=true]{hyperref}
\usepackage[noabbrev]{cleveref}
\usepackage{xcolor}
\usepackage{tabularx}
\usepackage{float}
\usepackage{pdflscape}
\usepackage{soul}
\usepackage{comment}
\usepackage{makecell}

\hypersetup{
    urlcolor=blue,
    citecolor=blue,
    linkcolor=black
    }
    
\usepackage[authoryear]{natbib}
\bibpunct[, ]{(}{)}{;}{a}{}{,}
\renewcommand\bibfont{\fontsize{10}{12}\selectfont}

\theoremstyle{plain}

\theoremstyle{definition}

\theoremstyle{remark}

\title{Assessment of the suitability of degradation models for the planning of CCTV inspections of sewer pipes}

\author{
\name{Fidae El Morer\thanks{CONTACT F. El Morer. Email: fidae.el.morer@tu-clausthal.de},  Stefan Wittek, Andreas Rausch}
\affil{Institute for Software and Systems Engineering, Technische Universit\"{a}t Clausthal, Clausthal-Zellerfeld, Germany}
}

\begin{document}
This is a preprint that has been submitted to the Urban Water Journal. This article is undergoing peer-review and is not accepted for publication. Feel free to contact the corresponding author and visit the repository containing the data and the results of this project. 

\maketitle
\begin{abstract}
The degradation of sewer pipes poses significant economical, environmental and health concerns. The maintenance of such assets requires structured plans to perform inspections, which are more efficient when structural and environmental features are considered along with the results of previous inspection reports. The development of such plans requires degradation models that can be based on statistical and machine learning methods. This work proposes a methodology to assess their suitability to plan inspections considering three dimensions: accuracy metrics, ability to produce long-term degradation curves and explainability. Results suggest that although ensemble models yield the highest accuracy, they are unable to infer the long-term degradation of the pipes, whereas the Logistic Regression offers a slightly less accurate model that is able to produce consistent degradation curves with a high explainability. A use case is presented to demonstrate this methodology and the efficiency of model-based planning compared to the current inspection plan.
\end{abstract}


\section{Introduction}
\subsection{Problem statement}
Physical assets in wastewater systems suffer from degradation over time, which translates into a constant loss from a financial and operational perspective. This deterioration can lead to damages that have health and environmental impacts due to exfiltrations that degrade the groundwater quality \citep{Bishop1998, Wolf2004}, sewer blockages that can lead to overflows \citep{Arthur2009, Rathnayake2019, Owolabi2022}, as well as interactions with other infrastructures such as roads \citep{Kuliczkowska2016, Dong2020}, among others.

A key component to prevent such impacts is an efficient operation and maintenance of sewer networks, which can be achieved with the definition of appropriate inspection strategies. Two main approaches to maintenance can be considered: reactive and proactive. Reactive techniques are based on intervening the assets only when they stop working, whereas proactive ones use preventive and predictive tools that anticipate the occurrence of failures \citep{Swanson2001}. \cite{Saegrov1999} suggests that proactive techniques have greater ``up-front" costs for the inspection, given the need of developing planning strategies to guide the decision making process, while greater ``follow-up" costs are derived from reactive strategies because failures might be already present in the assets when inspected. Therefore, a correct application and performance of proactive maintenance strategies can be more cost-efficient than the traditional reactive approach \citep{Fenner2000}.

The development of proactive maintenance strategies can also be seen as a planning system to prioritise what assets require to be inspected. Several authors have worked with different methodologies to establish prioritisation strategies for sewer asset maintenance. Many sewer network operators develop proactive planning strategies based on defining a fixed interval of years between subsequent inspections. In the case of Germany, the recommendations for the definition of inspection plans are set by the DIN EN 13508-1 \citep{din200313508-1}, but they are further developed by the states. In the case of the state of Nordrhein-Westfalen (Germany), the norm recommends to carry out the first inspection when the pipe is installed, another one after 10 years, and the rest of the inspections are performed every 15 years \citep{cremer2002wastewater}. 

This interval-based proactive or static planning  can be restrictive, given that robust or resilient pipes are being inspected when it is not strictly required, and critical or frail pipes are subject to inspections when the failure has already occurred. Furthermore, static planning does not take into consideration specific information about the structural or environmental features of the pipe, and it leaves out valuable information that arises from CCTV of inspections. Therefore, a dynamic planning or prioritisation system should be defined to take into account different factors that could cause the pipe to fail, as well as the information obtained from previous inspections, which shall be introduced as Dynamic Maintenance (DM). DM can be defined as a set of methods that use \textit{a priori} information such as the asset's age or the result of previous inspections to update the maintenance plan \citep{Bouvard2011}. To develop a DM plan for physical assets, a deterioration model is required. 

\subsection{Objectives}

Many statistical and machine learning-based degradation models have been presented over time, but most of them set their focus only on the accuracy metrics, without evaluating the ability of their models to produce long-term predictions of the deterioration of the assets. In order to develop DM plans, a long-term aging behaviour should be inferred from the results of the degradation model. Few examples can be found of degradation models where this property is assessed, but the results yield unrealistic behaviours where failure is never reached by the pipes \citep{Salman2012}, or the long-term simulations do not show a monotonic deterioration of the assets \citep{Caradot2018, bXianfei2020}, which is an inherent property of civil infrastructure systems where no maintenance is considered \citep{prakash2021toward}.

Additionally, the interpretability of the models should be taken into consideration. Although significant efforts have been made in recent years to elaborate methodologies that would allow machine learning models to be interpretable and go beyond the black-box paradigm \citep{ribeiro2016should}, the rationale behind the predictions cannot be understood and the internal logic is not transparent to the user or analyst \citep{guidotti2018survey, carvalho2019machine}. Given the lack of interpretability of black-box models, authors such as \cite{rudin2019stop} argue in favor of using inherently interpretable models in high-stakes decisions, so that the analyst or the user can have a transparent tool to decide whether to trust the predictions of the model or not.

Therefore, this work aims to provide a framework for the development of sewer deterioration models that goes beyond fitness or accuracy metrics. Two additional aspects should be considered to select a model for the planning of inspections, which include the generation of consistent long-term simulations that represent the probability of failure of the pipes along time, as well as its ability to produce interpretable and transparent results.

The main requirements that will be considered for the development of a satisfactory model are that a) it should accurately predict the condition of sewer pipes given a set of structural and environmental factors, b) the result of the simulation along time of single pipes must show a monotonic behavior, provided that the condition of the pipes cannot improve if no maintenance is considered, and c) the model should allow a certain level of interpretability in order to be able to explain the predictions conditioned on the inputs of the model. An additional contribution of this research paper is the inclusion of the length of the upstream network for every sewer pipe, which can be considered as a surrogate variable that accounts for the volume of water that flows through the pipes. 

The resulting model should be a useful tool for decision-makers and asset managers to schedule new CCTV inspections based on physical and environmental attributes of the sewer pipes and the result of previous inspection reports. Based on the probability of failure of each pipe, the decision-makers can elaborate sewer inspection plans with different levels of risk. To demonstrate the proposed methodology, a case study of a German urban area in the state of Nordrhein-Westfalen is presented.

The rest of this work is structured as follows: Section 2 is a literature review that covers the main contributions of previous works to the development of degradation models, focusing on statistical and machine learning classification models. In Section 3 we present the data for the use case and the methodology used to define the most suitable model. Section 4 covers the results of the comparison, as well as an example of the possible use of the resulting model. Section 5 presents the conclusions of this work.

\section{Related work}
Several authors suggested different consequence-based score systems that evaluate the effect of asset failures in the surrounding environment or in the operation of the sewer network itself. The higher the score given by this rating system, the greater the need to inspect and maintain a specific pipe. These methods use many factors such as the structural and physical characteristics of the pipes, the proximity of the assets to other critical infrastructures, or their importance within the network, and every variable has a weight assigned to it that reflects the relevance that it might have regarding the degradation process. As stated by their proponents, the main limitation of this approach is that it relies heavily on the subjectivity introduced by the developers of the model. These works include the ones presented by \cite{Arthur2009}, \cite{Baah2015}, \cite{Vladeanu2019} or \cite{Lee2021}.

Predictive models can overcome the drawback of the mentioned methods since no previous weights or influences need to be included in the system. Just like the aforementioned score systems, predictive models can include a myriad of factors that may cause the degradation of the assets. These predictive models are used to map some explanatory variables such as the physical attributes or the environmental information of the pipes to a scoring system that defines the condition of the pipe.

\subsection{Logistic Regression}
Logistic Regression (LR) models have been widely used in the literature to tackle the sewer pipe degradation problem. The works of authors such as \cite{Salman2012}, \cite{Sousa2014}, \cite{Kabir2018}, \cite{Laakso2019}, \cite{Robles-Velasco2021} or \cite{Fontecha2021} concluded that LR models are outperformed by more sophisticated machine learning methodologies, although the advantage shown by this type of statistical model is its transparency and the explainability through its coefficients. In order to look into the estimation of the coefficients of the LR model, \cite{Kabir2018} used a Bayesian approach that concluded that sewer age and length were the dominant drivers for the degradation of cementitious and clay pipes. As for the explainability on the predictions end,  \cite{Salman2012} proposed the use of LR for the development of degradation curves by simulating the life cycle of single pipes. The authors indicate that the degradation profiles show an unrealistic behavior for some materials, as their probability of failure in some cases reaches 50\% after 200 or 300 years.

\subsection{Random Forest}
Many authors have compared the use of Random Forests (RF) to classify both dichotomous and multiclass response variables that represent the condition of sewer pipes. The main proponents of this model are \cite{Harvey2014}, \cite{Laakso2019} and \cite{Hansen2020}. \cite{Caradot2018} compared the performance of different models to predict the condition of sewer pipes using three categories for the response variable. The authors performed a long-term simulation of the degradation behavior of individual pipes, noting that the prediction of the probability of failure decreased in certain periods of the simulations. They concluded that the interpretations that could arise from such a simulation could be misleading, as they would imply that the physical condition of pipes could improve along time even if no maintenance was carried out. Therefore, the authors recommend to use this approach only for \textit{ad-hoc} classification.

\subsection{Artificial Neural Networks}
Different architectures of Artificial Neural Networks (ANN) have been proposed by several authors to model the degradation behavior. Among these authors, we include \cite{Tran2006},  \cite{Khan2010}, \cite{Sousa2014}, \cite{Sousa2019}, \cite{aXianfei2020} and \cite{bXianfei2020}. From the mentioned works, only \cite{bXianfei2020} present deterioration curves for single pipes. The authors show several examples of long-term simulations for individual pipes and, as previously mentioned regarding the conclusions presented by \cite{Caradot2018}, the degradation curves that result from this model do not show a continuous deterioration of the pipes. 

\subsection{Other models}
Additional machine learning techniques have been proposed by other authors, although no degradation curves have been produced. Gradient Boosting models were used by \cite{aMalekMohammadi2020} and \cite{Fontecha2021}. The latter indicate that this model outperforms the rest of the prediction models subject to comparison, namely LR, RF and Decision Trees (DT). Support Vector Machines (SVM) were presented by \cite{Mashford2011},  \cite{Sousa2014} and \cite{Sousa2019}, concluding that although this algorithm showed a high potential in terms of predicting the condition of sewer pipes, ANNs yielded better results.

As shown in the previous paragraphs, the use of statistical and machine learning models has been widely explored and compared to predict the condition of sewer pipes with satisfactory results in terms of accuracy metrics, but there is still a gap in the assessment of the suitability of such tools for the application of degradation models that could be useful for the development of DM plans. In other words, it remains necessary to investigate the capacity of the proposed models to generate reliable and understandable outcomes, as well as consistent long-term simulations describing the deterioration of sewer pipes.

\Cref{tab:literature} shows a collection of the explanatory variables used by the mentioned authors in order to model the degradation of sewer pipes. For a more detailed review of the most influential factors in this field, we recommend the reviews conducted by \cite{bMalekMohammadi2020} and \cite{salihu2022towards}.

\begin{landscape}
\begin{table}
    \centering
    \caption{\label{tab:literature} List of reviewed research papers, including the explanatory factors and the techniques used to model the degradation of sewer pipes.}
    \footnotesize{
    \begin{tabularx}{\linewidth}{c|X|c}
        \hline
        \textbf{Authors} & \textbf{Explanatory variables} & \textbf{Models} \\
        \hline

        \cite{Mashford2011} & Diameter, age, road class, slope, Up/down invert elevation, material, grade, angle, soil corrosivity, sulfate soil/groundwater & Support Vector Machine \\
        &\\

        \cite{Salman2012} & Size, length, slope, age, depth, road class,
                                material, sewer function & Logistic Regression \\
        &\\
        \cite{Sousa2014} & Material, diameter, length, 
                            depth, slope, age, flow velocity & \makecell{Logistic Regression \\ Support Vector Machines \\ Artificial Neural Networks} \\
        &\\
        \cite{Sousa2019} & Material, diameter, sewer reaches, length, age, depth,
                            slope, flow velocity & \makecell{Support Vector Machines \\ Artificial Neural Networks} \\
        &\\
        \cite{Kabir2018} & Age, diameter, length, slope, depth, 
                            rim elevation, up and down invert & Bayesian Logistic Regression \\
        &\\
        \cite{Robles-Velasco2021} & Age, diameter, length, sewer function, soil type, 
                            shape, exposure to hydrogen sulphide, number of previous failures & Logistic Regression \\
        &\\
        \cite{Harvey2014} & Material, age, installation era, type of sewer, diameter, 
                            length, slope, slope change, up and down invert elevation, orientation change, depth,
                            road coverage, watermain breaks, land use, census tract & Random Forest Classifier \\
        &\\
        \cite{Laakso2019} & Slope, sewage flow, age, length, build year, coordinates, 
                            construction class, diameter, distance to trees, pipe type, material, depth, 
                            road class, stormwater pipe intersection, waterpipe intersection & \makecell{Logistic Regression \\ Random Forest} \\
        &\\
        \cite{Hansen2020} & Size, material, pipe function, land use, previous rehabilitations, 
                            distance to buildings, distance to trees, soil type, road class, 
                            position, slope, groundwater level & Random Forest \\
        &\\
        \cite{Caradot2018} & Age, material, shape, effluent type, district, length, 
                            width, depth, soil type, groundwater level, backwater,
                            distance to trees & Random Forest\\
        &\\
        \cite{Tran2006} & Size, age, depth, slope, tree-count, hydraulic condition, location, soil type, moisture index  & Artificial Neural Network \\
        &\\
        \cite{Khan2010} & Length, diameter, material, age, bedding material, depth & Artificial Neural Network \\
        &\\
        \cite{aXianfei2020} & Waste type, diameter, length, slope, water flow capacity, history of repairs,
                                pipe function, material, age & Artificial Neural Network \\
        &\\
        \cite{bXianfei2020} & Age, diameter, length, material, slope, average neighborhood LOF,
                                Waste type, Up/down stream depth, repair history, Up/down invert elevation, water flow capacity, pipe function & Artificial Neural Network \\
        &\\
        \cite{aMalekMohammadi2020} & Age, material, diameter, depth, slope, length, soil type, soils sulfate, 
                                    soil pH, groundwater level, soil hydraulic group, soil corrosivity, water flow & Gradient Boosting Trees \\
        &\\
        \cite{Fontecha2021} & Weather, population, previous failures, elevation, land use, slope, number of trees, 
         									gully pots, manholes, sewer pipes, streets, type of pipeline (local/main), pipe function (stormwater/sanitary) & \makecell{Logistic Regression \\ Decision Trees \\ Random Forest \\ Gradient Boosting Trees}\\
     \hline
    \end{tabularx}}
\end{table}
\end{landscape}

\section{Materials and methods}
\subsection{Data}
The use case that we present on this study is based on an urban area in the state of Nordrhein-Westfalen (Germany) with a population of around 25,000 inhabitants. The dataset is comprised by two main components, namely the physical and environmental attributes of the individual pipes, and the assessment of the condition of the sewer pipes carried out by experts based on CCTV inspections performed between the years 2000 and 2021. 

\begin{table}[H]
    \centering
    \begin{center}
    \caption{\label{tab:descriptive} Main statistics of the numerical predictors.}
    \small{
    \begin{tabular}{c|c|c|c|c}
        \hline
        \textbf{Variable} & \textbf{Min.} & \textbf{Max.} & \textbf{Mean} & \textbf{SD}\\
        \hline 
        Age & 0 & 74 & 30.199 & 16.705 \\
        Length & 1.43 & 175.27 & 34.082 & 16.190  \\
        Size & 100 & 2500 & 399.905 & 259.854 \\
        Depth & 0.394 & 7.22 & 2.316 & 0.924 \\
        Slope & -0.309 & 67.333 & 0.979 & 2.177\\
        Connection surface & 0.568 & 1263.832 & 170.077 & 113.118 \\
        Upstream length & 1.876 & 72009.122 & 1812.761 & 6205.549 \\
        Coordinates (X) & 0 & 1 & 0.348 & 0.1466 \\
        Coordinates (Y) & 0 & 1 & 0.547 & 0.219 \\
        \hline
    \end{tabular}}
    \end{center}
\end{table}
The database initially consisted of 12,832 inspections corresponding to 11,650 sewer pipe segments. Incomplete assessments or reports that contained missing values were left out of the analysis. As for the sewer pipes, house connections were not taken into consideration because although an inspection was carried out, no assessment on the condition was performed. The house connections account for 40.93\% of the inspections and 49.18\% of the pipes. Materials with less than 5 samples were excluded from the analysis, as no generalization could be drawn from such small groups. Finally, pipes that were given a very negative score despite being recently installed were dismissed, and the same goes for pipes that were installed 80 years prior to the inspection but were given the highest score in terms of condition (1.24\% of the inspections, 1.32\% of the pipes). These considerations resulted in a dataset with 6,279 inspections corresponding to 4,899 sewer pipe segments.

\subsubsection{Variable selection}
The list of variables considered for the development of the degradation model are shown in \cref{tab:variables}. Many of the variables taken into consideration such as the pipe length, the material or the average depth, have been considered previously by several authors. Additionally, this work proposes the use of the geographical coordinates of the sewer pipes' centroids as a surrogate variable for unobservable covariates such as groundwater fluctuations, soil compactation or interaction with infrastructures present in the surface, as suggested by \cite{Balekelayi2019}. To add further information about unobserved phenomena, this work includes the count and the length of upstream pipes, which can be considered a surrogate variable for the flow running through the pipes. Before training the models, the numerical variables have been properly scaled using a MinMax scaling. \Cref{tab:descriptive} shows the main descriptive statistics of the explanatory variables selected for this work. Note that the coordinates of the centroids of the pipes have been anonymized.

\begin{table}
    \centering
    \caption{\label{tab:variables} Input variables considered for the development of the model.}
    \footnotesize{
    \begin{tabularx}{\linewidth}{c|c|X}
        \hline
        \textbf{Variable name} & \textbf{Type} & \textbf{Description} \\
        \hline 
        Age & Numerical & Time elapsed between installation of the pipe and the inspection in years \\
        Length & Numerical & Length of the pipe segment in meters \\
        Size & Numerical & Height of the pipe segment in millimeters \\
        Depth & Numerical & Average depth of the pipe segment in meters \\
        Slope & Numerical & Slope of the pipe segment in percentage \\
        Connection surface & Numerical & Surface of the connection between the pipe and the manholes in squared meters \\
        Upstream length & Numerical & Length of the upstream pipes in meters \\
        Upstream pipes & Numerical & Count of upstream pipes \\
        Coordinates & Numerical & X and Y Geographical coordinates of the centroid of the pipe, expressed in degrees \\
        Material & Categorical & Material of the pipe segment \\
        Waste type & Categorical & Type of waste conducted by the pipe (wastewater, stormwater, mixed)\\           
        \hline
    \end{tabularx}}
\end{table}

\subsubsection{Response variable}
\begin{figure}[h]
\centering
\includegraphics[width=\textwidth]{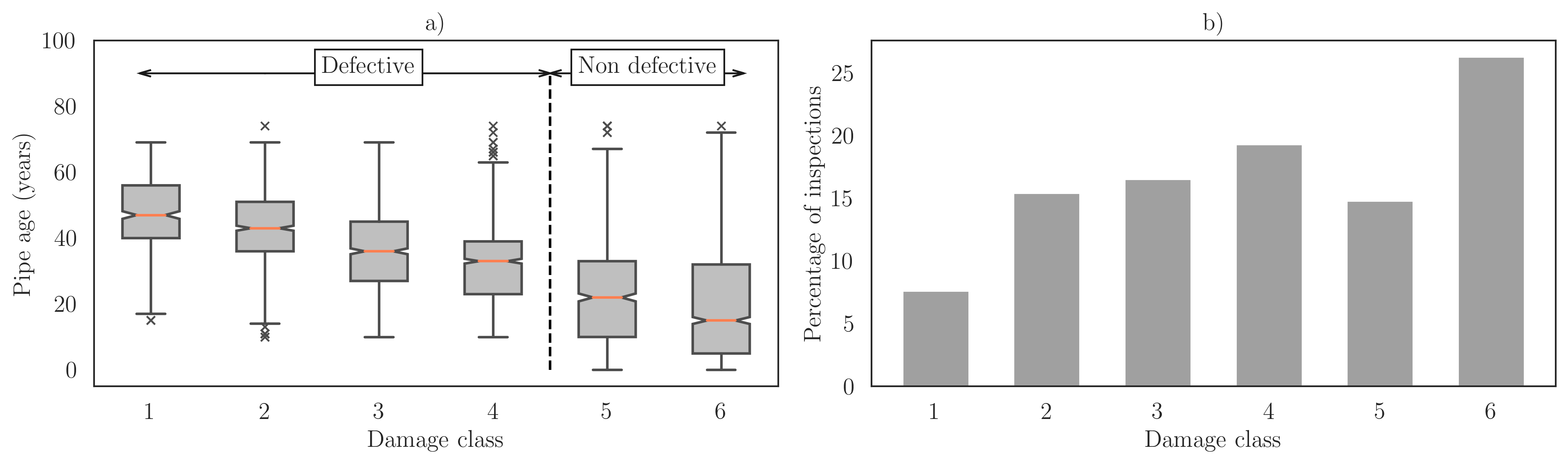}
\caption{a) Box plot showing the distribution of pipe age within damage classes. b) Count of inspections that fall within each damage class.}
\label{fig:substanz}
\end{figure}

The output variable is modelled based on the results of inspections carried out by experts. These inspections are performed according to the methodology provided by the ATV-M143-2 \citep{atv1999inspektion} and the DIN EN 13508-2 \citep{din200313508-2}, which state the guidelines for the interpretation and coding of damages using CCTV inspections. Based on these coding systems, the data provider uses an internal classification system from 1 to 6, where 6 indicates that the pipe is as good as new, and 1 means that the pipe should be replaced immediately. In order to simplify the modelling of such a variable, and to overcome the problem of class imbalance, the output has been binarized in such a way that classes 5 and 6 are considered non-defective, and the rest correspond to defective pipes. The binarization of the classes corresponding to different levels of structural or operational damage of sewer pipes can be found in previous works \citep{Salman2012, Harvey2014, aMalekMohammadi2020}. \Cref{fig:substanz} shows the result of the mentioned binarization, where it can be seen that there is a clear correlation between the pipe age and the damage class. Damage classes 5 and 6 account for 41\% of the observations as seen in the right-hand side of \cref{fig:substanz}. Considering these two categories under the same class (non-defective) helps to overcome the problem of class imbalance.

\begin{figure}[h]
\centering
\includegraphics[width=\textwidth]{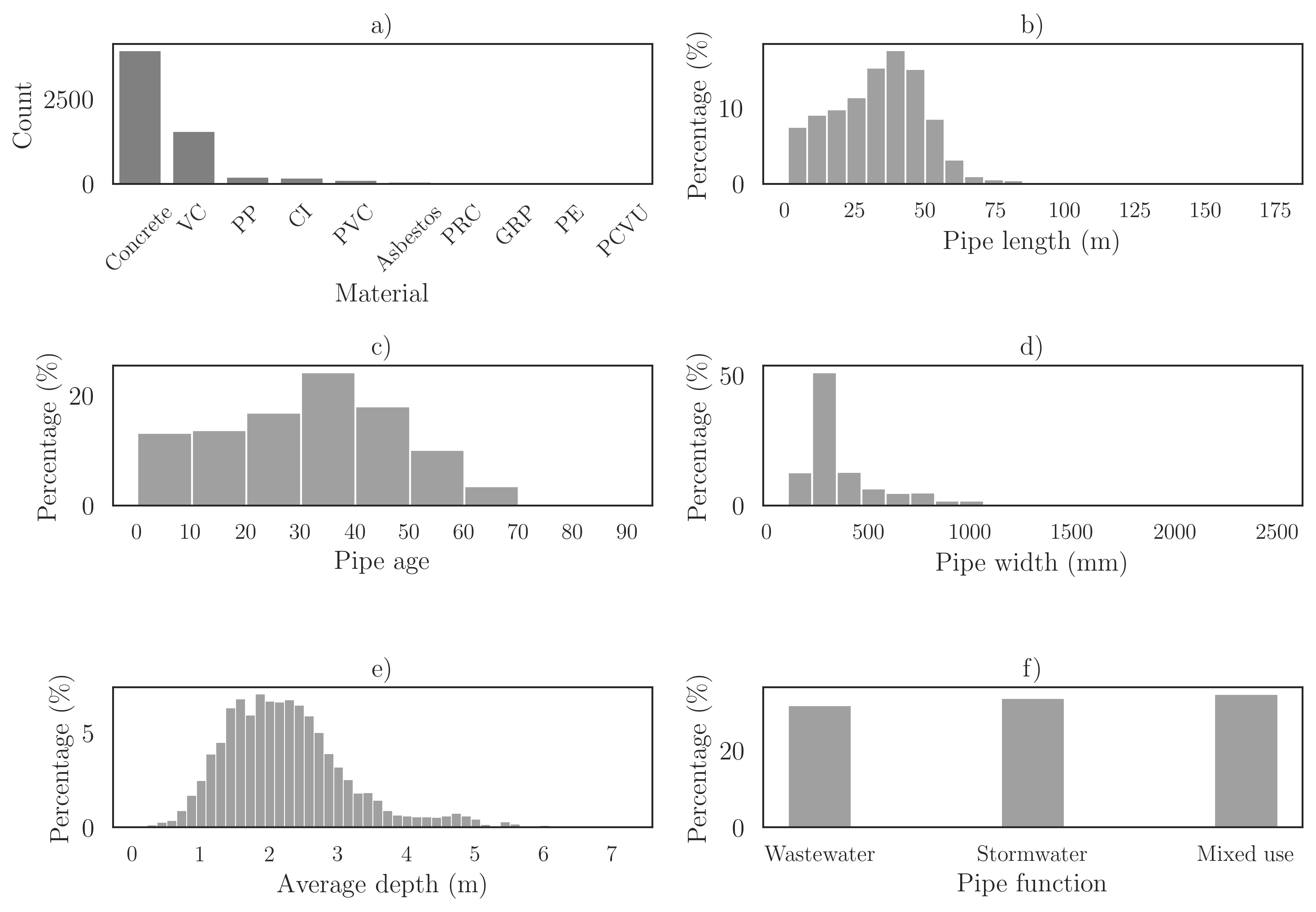}
\caption{Descriptive statistics of the main variables considered for the development of the model. (VC: Vitrified clay, PP: Ploypropylene, CI: Cast iron, PVC: Polyvinyl chloride, PRC: Polymer concrete, GRP: Glass reinforced plastic, PE: Polyethylene, PVCU: Unplasticised PVC)}
\label{fig:stats}
\end{figure}

The main descriptive statistics can be seen in \cref{fig:stats}. \Cref{fig:stats}(a) shows that the dataset mainly consists of concrete (63.53\%) and clay (25.20\%) pipes. As for the age of the pipes (\cref{fig:stats}c), 68.15\% of the samples were inspected before age 40, and only 3.69\% of the inspections correspond to pipes that were inspected after age 60, which implies a considerable bias towards pipes that were inspected shortly or moderately after their installation.

\subsection{Methodology}
\begin{figure}[H]
\centering
\includegraphics[scale=.8]{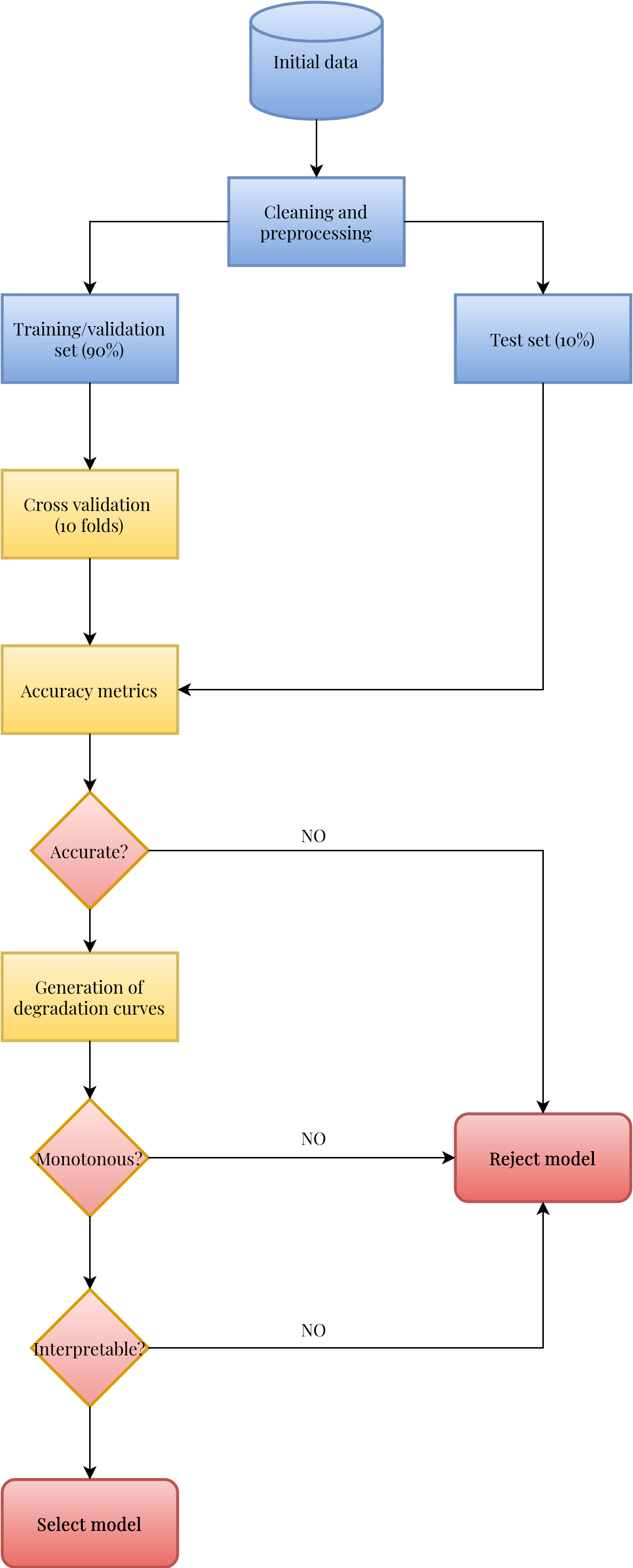}
\caption{Flowchart of the proposed methodology}
\label{fig:flowchart}
\end{figure}

As stated in previous sections of this work, the aim is to provide a predictive model that uses physical and environmental attributes of sewer pipes, as well as the results of prior assessments carried out after the performance of CCTV inspections, that is able to produce long-term degradation curves in order to develop DM strategies. To carry out such a task, two main assumptions will be made: a) the model should be able to accurately predict the condition of sewer pipes given the specified attributes and the response variable, and b) given that no maintenance, repairs or rehabilitation works are considered in the available inspections, the degradation curves that result from the simulation of the life cycle of the pipes should increase monotonically. Additionally the resulting model should be able to produce interpretable predictions based on the inputs. \Cref{fig:flowchart} shows a flowchart with the proposed methodology.

To achieve this goal, a set of statistical and machine learning models will be trained on the processed dataset. The performance of the models will be assessed under two criteria, namely the classification metrics specified on \cref{seq:metrics} and the temporal consistency of the degradation curves produced by the models.

\subsubsection{Models}
\label{seq:log_reg}
\subsubsection*{\textbf{Logistic Regression}}
The Logistic Regression (LR) is a statistical model that applies an inverse logit function to map a linear estimator to a binary outcome, having as a result the probability \(P \) of a sample \(x_{i} \) of belonging to the positive class (in the case of this work, the defective class), with a set of coefficients $\beta$. The linear estimator is composed by a matrix \(\textbf{X} \) that contains the values of the variables for each sample and a column vector $\beta$ which expresses the coefficients of said linear estimator. A link function $\sigma$ is applied on it, so the result of the estimator is constrained to the [0, 1] domain.

\begin{equation}\label{eq1}
    P(x_i; \beta ) = \frac{1}{1+e^{-\beta x_{i}}}
\end{equation}

LR models are inherently explainable, and they give information about the statistical power of the explanatory variables, as well as their effect on the response variable. Assuming that the model shows global significance, i.e. at least one of the coefficients is non-zero according to the result of the chi-square test, we must take into consideration the significance of the individual variables. The significance of the explanatory variables comes from applying a z-test to the standardized coefficients, and it shows the statistical power that a specific factor has to explain an event. 

Once a variable is considered significant, the coefficients can be interpreted by means of the Odds Ratio (OR). For an input variable \(j\) with a coefficient $\beta_j$, the OR is \(exp(\beta_j)\), and it can be interpreted as the odds that an outcome will occur given the presence of a specific factor, compared to the odds of the outcome occurring without that factor being present \citep{szumilas2010explaining}. For a variable with an OR$>$1, an increase in 1 unit of that factor will increase the probability of occurrence of the outcome. A formal definition and the interpretation of the results of the LR model can be found in \cite{demaris1995tutorial}.

\subsubsection*{\textbf{Decision Trees}}
Decision Trees (DT) are sequential models introduced by \cite{breiman2017classification} that perform a series of tests to find the optimal decision threshold for each variable in order to classify a sample. Each test is performed on a node, and each possible outcome of the test points out to a child node, where another test might be carried out. Subsequent tests are performed until a leaf is reached, which is a node without children \citep{kingsford2008decision}. 

The tests carried out in the nodes can be simplified as yes-no questions, which make the logical rules followed by the model easy to understand. Therefore, DTs can be considered inherently explainable models, as the logical process that they follow to produce results is explicit \citep{kotsiantis2013decision}.

\subsubsection*{\textbf{Random Forest Classifier}}
A Random Forest (RF) is an ensemble method introduced by \cite{breiman2001random} that combines the prediction results of several decision trees by means of averaging them. In terms of binary classification problems, RFs are constructed using a set of tree-structured predictors that cast a unit vote, and the output will fall into one of the two possible categories \{0, 1\}. For every input \(x_{i} \) from the collection of samples \(\textbf{X} \), the most popular predicted class $\hat{y_i}$ among the tree classifiers will be assigned.

RF models use Variable Importance (VI) as a measure of the relevance of an explanatory variable. A popular VI criterion is the Gini impurity, which is a metric used to decide the splits of the tree-structured predictors. Relevant predictors will have a higher decrease of the Gini impurity, and therefore, will have a higher VI \citep{archer2008empirical}. For a formal definition of this model, we recommend the works of \cite{breiman2001random} and \cite{biau2016random}.

\subsubsection*{\textbf{Extreme Gradient Boosting}}
Gradient Boosting (GB) machines are part of the boosting methods family. While classical ensemble techniques like RFs build predictions based on weak estimators, boosting methods add new models to the ensemble sequentially \citep{natekin2013gradient}. In this sense, the model initially proposed by \cite{friedman2001greedy} aims at sequentially building new base-learners to be maximally correlated with the negative gradient of the loss function. The GB model used in this work is based on the XGBoost library, developed by \cite{chen2016xgboost}, which presents an efficient and scalable implementation of this technique.

Similarly to RFs, GBs can give a measure of the relevance of the inputs to generate the output variable. This is done through the gain, which is a metric used to optimize the splits of the boosted trees. A variable that increases the gain is more decisive for the development of the model, and therefore, it is more relevant to explain the output.

\subsubsection*{\textbf{Support Vector Machine}}
Support Vector Machines (SVM) were initially introduced by \cite{boser1992training} as an algorithm to find the optimal decision boundary between classes. For the two-class discrimination problem, SVMs determine a separating hyperplane (or decision boundary) in a high-dimensional space, relying on maximizing the margin or minimal distance between the hyperplane and the closest data points to it \citep{mammone2009support}. An advantage presented by such a model is the possibility of selecting different kernels, which are mathematical devices that project the data samples from a low-dimensional space to a space of higher dimension. This transformation allows the data to become separable in the higher space by means of the aforementioned hyperplane \citep{noble2006support}.

\subsubsection*{\textbf{Artificial Neural Networks}}
Artificial Neural Networks (ANN) are a set of models that correspond to the family of deep learning techniques and are widely used for pattern recognition problems. The structure of ANNs is composed of an input layer where the features of the data samples are introduced, a set of hidden layers, and an output layer, where the target value is approximated. These layers are made of neurons, which are computational or processing units that apply linear or non-linear transformations (activation functions) to the information coming from previous layers during the feedforward step. The optimization of the parameters of the ANNs comes from the backpropagation step, which takes into consideration the error of the prediction during the feedforward step, and updates the values of the parameters to yield a better estimate of the outputs given the inputs. For a better understanding of this type of models, we recommend the work of \cite{krogh2008artificial}, and for a formal definition of neural networks, we suggest \cite{jain1996artificial}.

In the context of this work, an ANN with 2 hidden layers consisting of 100-50 neurons respectively with a Rectified Linear Unit (ReLU) as the activation function was used. The output layer consists of a single neuron with a sigmoid activation function, since the aim of the model is to discriminate between two classes. 

\label{seq:metrics}
\subsubsection{Model quality metrics}
Several classification metrics have been used to compare the performance of the models. Given a binary outcome, 4 possible predictions can arise after training a model, namely true positives (TP), true negatives (TN), false positives (FP) and false negatives (FN), assuming that in this context, a positive value would represent a defective pipe.

The accuracy (\cref{eq2}) represents the proportion of correct predictions with respect to the sample size. It is a good estimator of the performance of a model, but it does not give information about the bias of the model in terms of leaning towards FNs or FPs.

\begin{equation}\label{eq2}
    Accuracy = \frac{TN+TP}{TN+TP+FN+FP}
\end{equation}

The precision (\cref{eq3}) or positive predictive value is the proportion of TPs over the total positive predictions. That is, in this context, the precision would represent the rate of samples that were correctly predicted as damaged with respect to the total amount of samples that were considered damaged by the model.

\begin{equation}\label{eq3}
    Precision = \frac{TP}{TP+FP}
\end{equation}

The recall (\cref{eq4}) or true positive rate shows the proportion of TPs with respect to the known positives. In the context of this work, it would represent the rate of observations that were considered damaged (positive) with respect to all the samples that were actually damaged.

\begin{equation}\label{eq4}
    Recall = \frac{TP}{TP+FN}
\end{equation}

Finally, the Area Under the Curve (AUC) is used as a metric for the performance of the models. This metric comes from the Receiver Operating Characteristic (ROC) curve, which shows, for different thresholds, the relationship between the TP ratio and the FP ratio. A perfect classifier would have a ROC curve that reaches a value of 1 for the TP ratio and 0 for the FP ratio simultaneously, and therefore, the AUC would have a value of 1. For a more detailed description of the presented metrics, we suggest the review presented by \cite{lever2016classification}.

\label{seq:mono}
\subsubsection{Monotonicity}
As stated in previous sections, degradation curves are expected to increase monotonously with respect to time, given that no maintenance tasks are considered. To check whether this condition is fulfilled by the tested models, a simple algorithm will be ran to compute if, at a certain age \(t\), the probability of being defective \(P(x_{t})\) is higher than the same probability one year before. If \(P(x_{t}) < P(x_{t-1})\), the behavior will not be considered monotonous.

\section{Results and discussion}
\subsection{Performance metrics}
\begin{figure}[h]
\centering
\includegraphics[width=\textwidth]{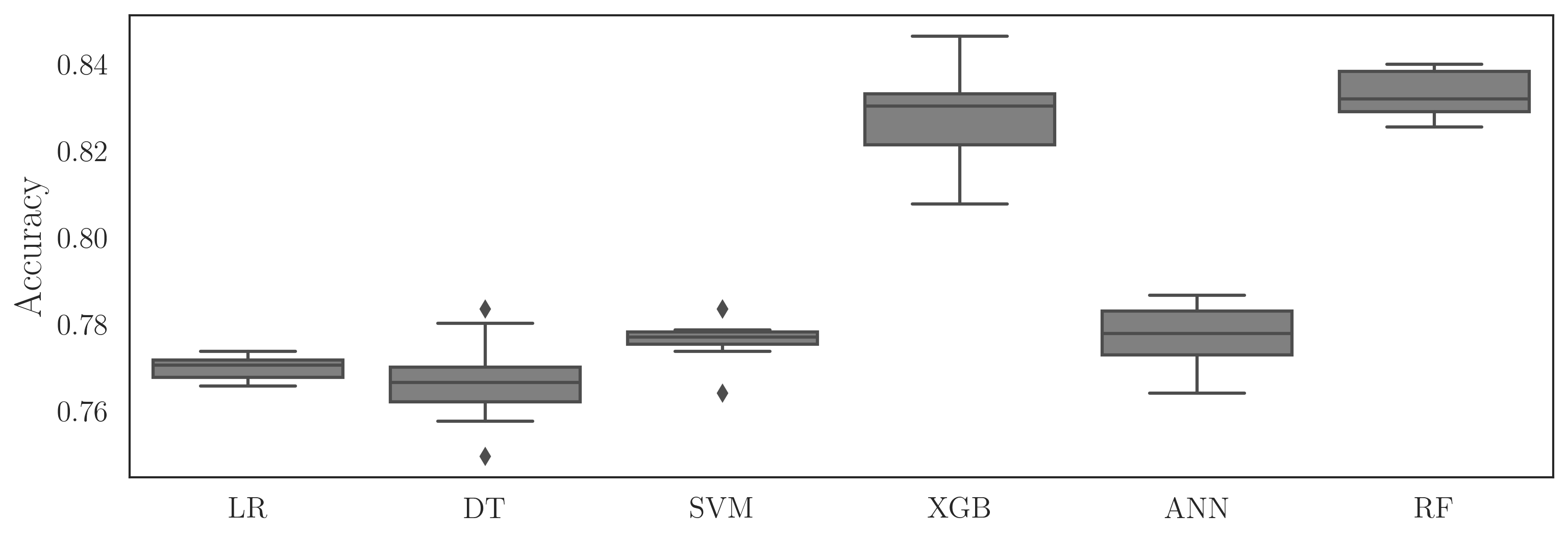}
\caption{Cross-validated accuracy scores of the models on the held-out test set.}
\label{fig:cross}
\end{figure}

The performance of the proposed models is compared using cross-validation. 90\% of the data is selected for training and validation purposes, and 10\% is held-out to assess their ability to generalize to unseen samples. The cross-validation is applied on the first batch (training and validation set) so that 70\% of the samples are used for training and 30\% are for validation. This process is repeated across 10 folds, and in every iteration, the performance metrics are calculated on the held-out (test) set. The cross-validation (\cref{fig:cross}) shows that there is a significant difference between the performance of the ensemble models (XGB and RF) with respect to the rest of the tested techniques, as suggested by authors such as \cite{Laakso2019} or \cite{Fontecha2021}. The mentioned models show a higher accuracy, but also a higher variance than other models. No significant difference can be seen on average between the ANN and the SVM models, although the SVM shows a much lower variance. The LR shows a high robustness in its predictions, but its accuracy is lower than that shown by the SVM and the ANN. Finally, the DT shows the lowest accuracy on average, and it presents a variance comparable to that of the ensemble models or the ANN.

A similar pattern can be observed regarding the rest of the performance metrics. RF shows a higher recall, precision and AUC than the rest of the models, followed closely by the XGB. This means that not only the ensemble models outperform the rest in terms of accuracy, but they also provide more reliable predictions, given the balance between the rates of FNs and FPs. SVM shows a similar recall to the one yielded by the ensemble methods, but it has the lowest precision, which means that the model is biased towards predicting more FPs than FNs. The result given by the SVM implies that the model would be prone to suggest that a pipe is defective when it is not. As seen in \cref{fig:cross}, the LR shows a comparable accuracy to the DT, the SVM or the ANN, although it outperforms the last 2 models in terms of precision. The LR model also shows a lower variance in the performance metrics, thus rendering this model more robust in terms of its predictions.

Despite the inherent difficulty to perform comparisons across different studies (different target values, uncertainty of the pipe condition inspections and metrics, different input variables, etc.), the results that have been obtained in this research work are consistent with the literature review. \cite{Laakso2019} and \cite{Fontecha2021} show that ensemble models such as XGB or RF outperform simpler models like the LR.

\begin{table}
    \centering
    \begin{center}
    \caption{\label{tab:model-results} Comparison of the average performance metrics (and standard deviation) of the tested models. Random Forest yields the best results for all the metrics.}
    \small{
    \begin{tabular}{c|c|c|c|c}
        \hline
        \textbf{Model} & \textbf{Accuracy} & \textbf{Recall} & \textbf{Precision} & \textbf{AUC} \\
        \hline 
        LR & 76.995 (0.254) & 81.921 (0.595) & 78.721 (0.321) & 0.849 ($9.59\times10^{-4}$) \\
        DT & 76.704 (0.998) & 80.904 (1.229) & 78.908 (1.047) & 0.759 ($1.045\times10^{-2}$) \\
        SVM & 77.657 (0.544) & 89.039 (0.921) & 76.014 (0.652) & 0.863 ($1.926\times10^{-3}$) \\
        XGB & 82.827 (1.063) & 90.734 (1.317) & 81.382 (0.873) & 0.898 ($4.995\times10^{-3}$) \\
        ANN & 77.689 (0.763) & 86.779 (3.244) & 77.185 (1.893) & 0.869 ($6.576\times10^{-3}$) \\
        RF & 83.311 (0.533) & 91.356 (1.058) & 81.666 (1.023) & 0.911 ($3.455\times10^{-3}$) \\
    \end{tabular}}
    \end{center}
\end{table}

\subsection{Degradation curves}
\begin{figure}[h]
\centering
\includegraphics[width=\textwidth]{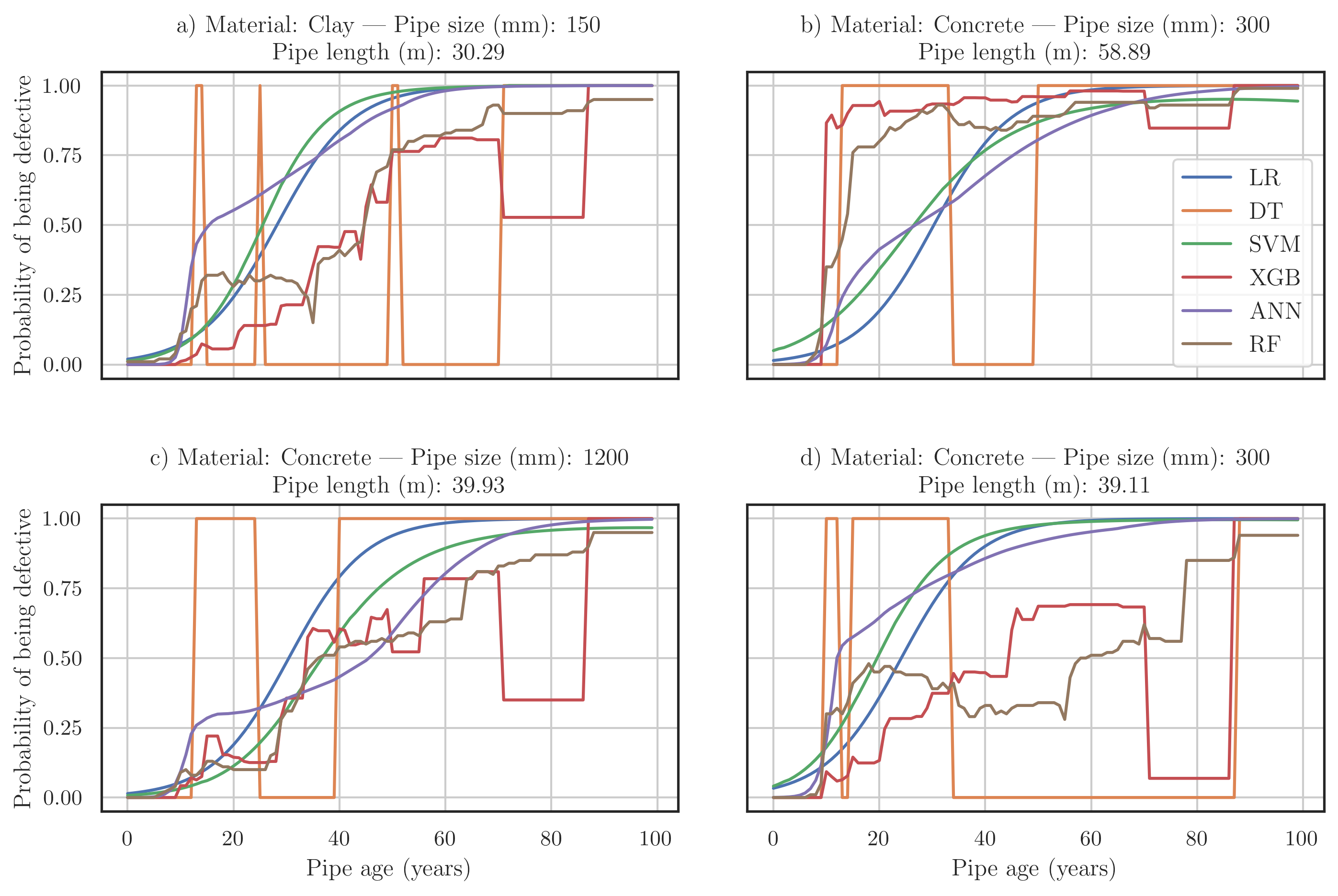}
\caption{Degradation curves showing the probability of failure of four different sewer pipes according to the trained models.}
\label{fig:models}
\end{figure}

To illustrate the differences between models in terms of their capability of generating degradation curves, \cref{fig:models} shows the results of the simulation of 100 years of 4 different pipes. These pipes are selected after carrying out the monotonicity test (\cref{tab:monotonous}), and represent the samples with the highest amount of decreases in terms of the probability of failure along time. \Cref{fig:models}(a) shows the sewer pipe where DTs yield more shifts, \cref{fig:models}(b) represents the same for the SVM model, \cref{fig:models}(c) for the XGB and \cref{fig:models}(d) for the RF. 

The DT model only captures the extreme probability values, i.e. 1 and 0, which makes it an unsuitable for the prediction of probabilities, as it only produces binary values, and they are not consistent with the aging behavior.

The ensemble models, i.e. XGB and RF, show similar behaviors as the probability of failure increases along time, but both of them fail to show a monotonic degradation curve. XGB shows a spiky curve with a sudden drop in the probability of failure after ~70 years, and after a short period it rises up again to reach a probability of failure of 100\%. As for the RF, the probability of failure only reaches 100\% in the case of \cref{fig:models}(b), and even if it shows a general upward trend, the model suggests an improvement of the condition of the pipes at different ages. This result is in line with the findings presented by \cite{Caradot2018}, where the authors indicate that this long-term forecast could be misleading, since it would be suggesting that the pipe will improve its structural and operational condition along time.

LR and SVM models show a similar pattern in the predicted degradation behavior. Both models generate S shaped curves that show a smooth increase in the probability of failure, although the predictions produced by the SVM do not always reach a probability of failure of 100\%, and the curve showed in \cref{fig:models}(b) shows a decay in the degradation rate, which would imply an improvement in the condition of the asset.

As seen on \cref{tab:monotonous}, ANNs yield monotonic degradation curves for all the simulations, although the predicted behavior is more irregular than the one showed by the LR or the SVM.

\begin{table}
    \centering
    \begin{center}
    \caption{\label{tab:monotonous} Metrics of the models in terms of the monotonicity of the degradation curves compared across every sample in the dataset.}
    \small{
    \begin{tabular}{c|c|c|c}
        \hline
        \textbf{Model} & \textbf{Count} & \textbf{Mean} & \textbf{Max} \\
        \hline
        LR & 0 & 0 & 0 \\
        DT & 8,224 & 1.33 & 4 \\
        SVM & 16,504 & 2.67 & 14 \\
        XGB & 44,033 & 7.12 & 15 \\
        ANN & 0 & 0 & 0 \\
        RF & 54,593 & 8.82 & 24  \\
        \hline
    \end{tabular}}
    \end{center}
\end{table}

\subsection{Interpretability}
Among the two models that produce degradation curves that show a monotonic increase of the probability of failure, the LR is the only one that yields an interpretable result based on the coefficients of its linear estimator. By means of these coefficients, it is possible to know what is the size of the effect of the input variables with respect to the output, as well as its sign and its statistical significance. \Cref{tab:lr-results} shows the coefficients obtained from training the LR model. 

As stated in \cref{seq:log_reg}, the interpretation of the LR can be done by analyzing its coefficients and the ORs. The results of the analysis show that, as reported in previous studies, pipe age and structural features such as its length or size are highly significant factors when it comes to sewer pipe degradation. Observing the OR of the pipe age, it can be seen that an increase of 1 year of age rises the chances of the pipe being defective by a ratio of 1.095. 

The function of the pipe only appears to be significant when they transport stormwater. Its negative coefficient and its OR $<$1 indicate that mixed use and sewage pipes are more prone to degradation than stormwater pipes. This result is highly dependant on the maintenance strategies carried out by the water utility. As seen in \cite{DAVIES200173} and \cite{Baur2002}, the degradation of mixed use sewers is lower due to higher engineering, construction and maintenance efforts, whereas the results obtained by \cite{Salman2012} show that sanitary pipes are more resilient to deterioration. On the contrary, the size of the pipe, represented in this analysis by the height, shows a statistically significant negative effect on the outcome, which suggests that bigger pipes are more resilient, which is in line with the conclusions of authors such as \cite{Salman2012} or \cite{Bakry2016}.

The length of the pipes slightly increases the probability of failure. Authors such as \cite{ana2009}, \cite{Khan2010} or \cite{Laakso2019} explain this effect arguing that longer pipes have more joints, which are vulnerable to failure, and are more exposed to structural defects such as bending. The length of the upstream pipes has a similar effect on the outcome, showing that the probability of failure could be correlated with the volume of water flowing through the pipes, considering that downstream pipes will receive a higher volume. As stated previously, this result depends on the particular characteristics of the studied network and the asset management strategies performed by the water utility, and it should not be confused with the effect of the flow rate on the degradation of the pipes. According to authors like \cite{Tran2006} or \cite{Salman2012}, steep slopes cause higher flow rates, which lead to higher deterioration rates, whereas lower slopes can cause sedimentation due to the low velocity of the water \citep{Laakso2019}. Given the lack of statistical significance of the slope in the presented experiment, no conclusion about the correlation between flow rate and sewer degradation can be extracted for this particular use case.

\begin{table}

\tbl{Coefficient estimates, significance and Odds Ratios of the variables used in the Logistic Regression model.}
{\begin{tabular}{lcccccccc} \toprule
 & \multicolumn{5}{l}{Coefficients} & \multicolumn{3}{l}{Odds Ratio} \\ \cmidrule{2-9}
 Variable & Estimate & Std. error & z value & P($>$$|z|$) & Significance & Estimate & CI 2.50\% & CI 97.50 \%\\ \midrule
        Pipe age & 9.141$\times10^{-2}$ & 2.795$\times10^{-3}$ & 32.706 & $<$$2\times10^{-16}$	& *** &	1.095 & 1.089 & 1.101 \\
        Upstream length & $2.818\times10^{-2}$ & $6.367\times10^{-3}$ & 4.426 &	$7.391\times10^{-6}$ & *** & 1.028 & 1.016 & 1.041 \\
        Pipe length & $9.915\times10^{-3}$ & $3.212\times10^{-3}$ & 3.087 &	$2.024\times10^{-3}$ & ** &	1.009 &	1.003 &	1.016\\
        Pipe size  & -1.527 & 0.183 & -8.310 & $<$ $2\times10^{-16}$ &	*** & 0.217 & 0.151\ &	0.312\\
        Connection surface  & $1.284\times10^{-3}$ & $5.392\times10^{-4}$ &	2.381 &	$1.775\times10^{-2}$ &	* &	1.001 &	1.000 &	1.002 \\
        Depth & $3.824\times10^{-2}$ & $5.955\times10^{-2}$ & 0.642	& 0.521 & & 1.038 &	0.924 &	1.167 \\
        Slope & $5.231\times10^{-3}$ & $8.681\times10^{-3}$ & 0.603 & 0.546 & & 1.005 &	0.987 & 1.024 \\
        Material & & & & & & & \\
        \hspace{3mm}Asbestos & 0.643 & 0.534 & 1.204 & 0.228 &  & 1.903 & 0.683 &	5.576 \\
        \hspace{3mm}Concrete & 0.216 & 0.412 & 0.524 & 0.600 & & 1.241 & 0.562 & 2.851 \\
        \hspace{3mm}CI & -4.469 & 0.481 & -9.275 & $<$ $2\times10^{-16}$	& *** &	$1.146\times10^{-2}$ & $4.484\times10^{-3}$ & $2.977\times10^{-2}$ \\
        \hspace{3mm}PE & -13.882 & $2.854\times10^{2}$ & $-4.924\times10^{-2}$ & 0.961 & & $9.372\times10^{-7}$ & $4.523\times10^{-135}$ & $3.787\times10^{-115}$  \\
        \hspace{3mm}PP & -2.043 & 0.602 & -3.393 & $6.983\times10^{-4}$ & *** & 0.129 & $3.828\times10^{-2}$ & 0.411\\
        \hspace{3mm}PRC & -13.623 & 19.703 & $-6.928\times10^{-2}$ & 0.9448 & & $1.215\times10^{-6}$ & $5.175\times10^{-102}$ & $8.901\times10^{-89}$ \\
        \hspace{3mm}PVC & -0.845 & 0.476 & -1.773 & $7.623\times10^{-2}$ & . &	0.429 & 	0.171 & 1.110 \\
        \hspace{3mm}PVCU & -0.549 & 0.739 & -0.743 & 0.457 &	 & 0.577 & 0.126 & 2.356 \\
        \hspace{3mm}Clay & -0.199 & 0.427 &	-0.468 & 0.640 & & 0.818 & 0.361 & 1.934 \\
        Sewer type & & & & & & & \\
        \hspace{3mm}Combined sewer & $8.704\times10^{-2}$ & $9.129\times10^{-2}$ & 0.952 & 0.341 & & 1.091 & 0.912 &	1.305 \\
        \hspace{3mm}Stormwater & -0.359	& 0.126	& -2.858	& $4.271\times10^{-3}$ & ** & 0.698 & 0.545 &	0.893 \\
        Coordinates & & & & & & & \\
        \hspace{3mm}X & -10.2 & 1.542 & -6.611 & $3.812\times10^{-11}$	& *** & $3.741\times10^{-5}$ & $1.803\times10^{-6}$ & $7.620\times10^{-4}$ \\
        \hspace{3mm}Y & 1.432 & 0.226 & 6.435 & $1.241\times10^{-10}$ &	*** & 4.187 & 2.710 &	6.487\\ \bottomrule
\end{tabular}}
\label{tab:lr-results}
\tabnote{‘***’: p$<$0.001; ‘**’: p$<$0.01; ‘*’: p$<$0.05; ‘.’: p$<$ 0.1}
\end{table}

\begin{figure}[h]
\centering
\includegraphics[width=\textwidth]{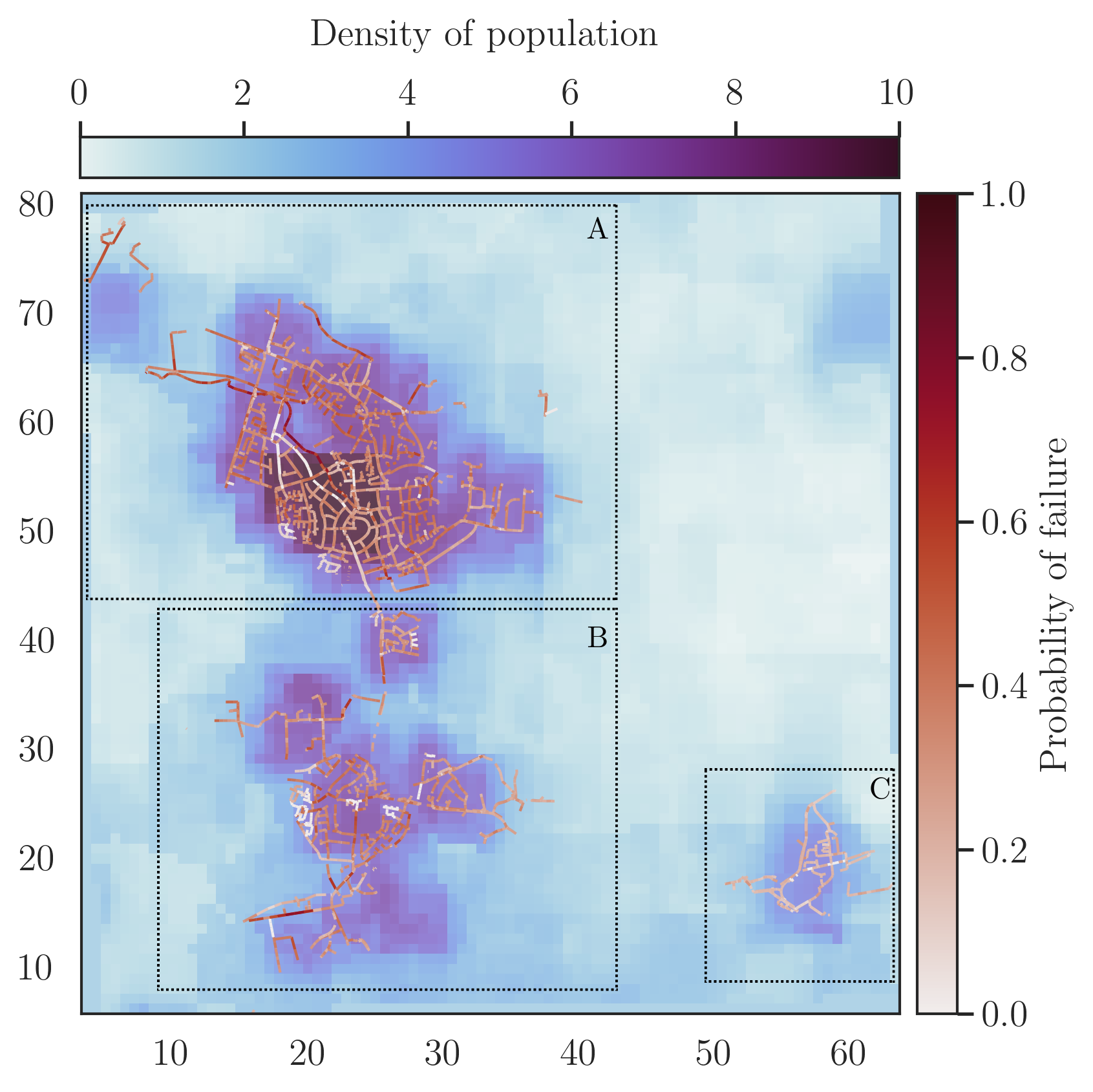}
\caption{The map shows the density of population of the area under study, represented by the approximated count of inhabitants in each pixel. The color of the pipes represent the probability of failure at age 20 of each pipe. Pipes located in the upper left part of figure show a higher probability of failure than the ones located in the opposite side of the area under study. Population density map provided by WorldPop \citep{tatem2017worldpop}.}
\label{fig:map}
\end{figure}

Finally, the X and Y coordinates of the centroid of the pipe show opposite effects on the response variable. The OR of the X coordinates suggests that an increase in 1 unit of this variable lowers the chances of the pipe being defective (OR $<$ 1), meaning that pipes that are situated in eastern areas of the studied area show a slower degradation rate. The coefficient related to the Y coordinates indicates exactly the opposite: pipes situated in the North are more prone to failure (OR $>$ 1). This difference can be better explained by looking at \cref{fig:map}, where it can be clearly seen that region A, which lays in the Northwest of the studied area, has a higher density of population, and therefore, a higher density of sewer pipes and a higher volume of water. On the contrary, Region C (Southwest) is less populated than its counterparts, and it has a less complex sewer network. For the same simulated age, this area shows lower probabilities of failure, confirming the intuition behind the size and the sign of the coefficients regarding the coordinates of the centroids of the pipes and the length of the upstream pipes.

\subsection{Current inspection strategy vs. model-based strategy}

Once the best option is selected among the proposed models, a comparison can be made between the current inspection plan and the one that can be drawn from exploiting the model. The advantage of using the proposed model is that it provides flexibility in setting probability thresholds. By adjusting the threshold, the planner can determine the acceptable level of risk and allocate inspection resources accordingly. This flexibility allows for a more efficient inspection plan, focusing resources on pipes with higher probabilities of failure.

\begin{figure}[H]
\centering
\includegraphics[width=\textwidth]{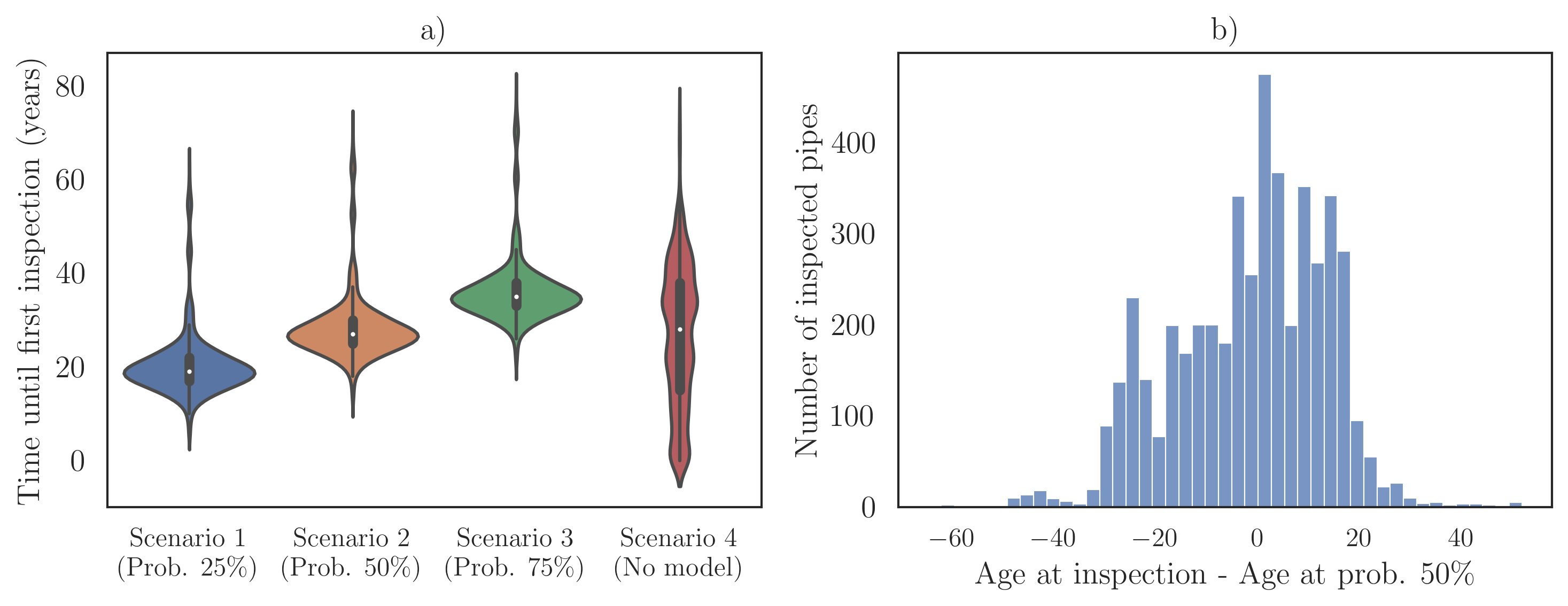}
\caption{a) Comparison of alternative scenarios considering different probability thresholds with the current scenario b) Difference of pipe ages at inspection between the current strategy and the results of a model with a probability threshold of 50\%.}
\label{fig:simulations}
\end{figure}

\Cref{fig:simulations}a shows a comparison of 4 different scenarios where 3 possible probability thresholds are defined against a scenario where no model is used. For example, in Scenario 1, a conservative threshold is set, resulting in a large number of pipes being inspected. This approach prioritizes safety but may lead to unnecessary inspections and increased costs. In Scenario 2, a moderate threshold is used, reducing the number of inspections compared to Scenario 1 while still maintaining an acceptable level of risk. Scenario 3 represents a more risk-tolerant approach with a higher threshold, resulting in even fewer inspections.

When comparing Scenario 4 (no model) and Scenario 2 (with a probability threshold of 50\%), approximately half of the network is inspected after around 27 years. By subtracting the predicted failure age of the pipes from the actual age of inspection (as shown in Figure 2b), we obtain a distribution where some pipes are inspected before the predicted cutoff point (negative side) and others are inspected later than required (positive side).

In this case, according to the model and the selected probability threshold, 49.11\% of the pipes are inspected later than required, which could lead to higher maintenance and reparation costs. A more restrictive strategy such as the one proposed in Scenario 1 would lead to a proportion of 68.23\% of the pipes inspected too late, and Scenario 3 would result in a rate of 28.71\% of this quantity. Therefore, to optimize the operation and maintenance of the sewer network, the decision boundary (probability threshold) needs to be adjusted accordingly. This adjustment should take into account the needs and resources of the managing authority to strike a balance between timely inspections and cost-effectiveness.

\section{Conclusions}

This work presented a comparison of different statistical and machine learning methods to assess their suitability to tackle the problem of modelling the degradation of sewer pipes. The analysis has been carried out considering three main elements, namely the accuracy of the models, their ability to produce consistent long-term simulations based on the probability of failure of single pipes, and their interpretability.

The results showed that ensemble methods such as Random Forests or Gradient Boosting Trees yield the best results in terms of accuracy metrics, but their long-term simulations do not produce monotonous degradation curves, which implies that they cannot be used to develop reliable dynamic maintenance plans in the presented scenario. Support Vector Machines and Artificial Neural Networks show similar accuracy metrics, but the former is not able to generate coherent long-term simulations, and the latter lacks the interpretability that was seeked during the presentation of the requirements of this work. The Logistic Regression showed slightly less accurate results, but it produced degradation curves that fulfilled the monotonicity requirement, and is inherently explainable by means of its coefficients, rendering it the most suitable model for the development of dynamic inspection plans for the presented use case.

After obtaining these findings, a simulation was conducted to compare the existing situation (without a model) with three alternative scenarios employing various thresholds for the probability of failure of single pipes. This simulation demonstrated the effectiveness of a data-driven model to prevent a high proportion of pipes of the network from being inspected later than required.

This study has provided a framework to assess different statistical and machine learning models for creating inspection plans that consider long-term failure simulations and model interpretability. However, further research is needed to make the methodology more reliable. This can be achieved by analyzing larger datasets that include more variables affecting sewer pipe deterioration and comparing the costs of different inspection plans to the current scenario.

\section*{Acknowledgements}
This research was funded in part by the German Federal Ministry of Education and Research (BMBF) under the project KIKI (grant number 02WDG1594A). The authors would like to acknowledge the support and collaboration of August-Wilhelm Scheer Institut f{\"u}r digitale Produkte und Prozesse GmbH, IBAK Helmut Hunger GmbH \& Co. KG, Eurawasser GmbH \& Co. KG, AHT AquaGemini GmbH and Entsorgungsverband Saar.

\section*{Data availability}
The data used in this study is available upon request from the corresponding author or can be accessed through the following GitHub repository: \url{https://github.com/Fidaeic/sewer-pred}.

\bibliography{Sewer_pred}

\end{document}